\documentclass[journal]{IEEEtran}
\usepackage[numbers]{natbib}

\usepackage[utf8]{inputenc} 
\usepackage[T1]{fontenc}    
\usepackage{hyperref}       
\usepackage{url}            
\usepackage{booktabs}       
\usepackage{amsfonts}       
\usepackage{nicefrac}       
\usepackage{microtype}      
\usepackage{xcolor}         
\usepackage{wrapfig}
\usepackage{multirow}
\usepackage{graphicx}
\usepackage[table]{xcolor}
\usepackage{comment}
\usepackage{enumitem}
\usepackage{tcolorbox}
\title{\texttt{TeleResilienceBench}: Quantifying Resilience for LLM Reasoning in Telecommunications}

\author{%
  Pranshav Gajjar$^1$,  Emmanuel Ojo$^1$, and Vijay K Shah$^1$\\
  $^1$NextG Wireless Lab,\\
  North Carolina State University\\
}

\begin{document}

\maketitle

\begin{abstract}

Deploying large language models in telecommunications requires more than task accuracy. In realistic workflows, a model may inherit partially completed reasoning from a prior step, an upstream agent, or its own earlier generation, and must continue that reasoning even when it is already going wrong. We introduce \texttt{TeleResilienceBench}, a benchmark that quantifies this capability, which we term \emph{reasoning resilience}, across seven telecom sub-domains drawn from the GSMA Open-Telco LLM suite. Instances are constructed by collecting failures from a weak generator model, truncating the flawed reasoning trace at its midpoint, and asking a target model to continue and correct it. We propose the Correct Flip Rate (CFR) as a direct measure of successful recovery and evaluate eight models spanning the Qwen3.5, Gemma4, and Nemotron-3 families. Our results show that even the strongest model achieves a macro-average CFR of only $29.1\%$, and scale does not reliably improve resilience within families. Nemotron-3-nano 4b outperforms all Qwen3.5 variants including the 27b model and leads the auxiliary TeleMath numerical evaluation at $23.4\%$ CR\%, offering the best resilience-to-cost ratio in the set. A difficulty-stratified analysis further reveals that existing telecom benchmark difficulty labels reflect factual specificity rather than reasoning depth, suggesting that current evaluations measure knowledge coverage more than reasoning ability.

\end{abstract}


\section{Introduction}



Large language models (LLMs) are rapidly shifting from general-purpose assistants to domain-specific components for technical decision support.
In telecommunications, this transition is being actively driven by the AI-RAN paradigm, championed by industry leaders including NVIDIA and
Nokia, which envisions LLMs as native operational components embedded directly within radio access network
infrastructure~\citep{nvidia2025nokiaairan, nvidia2026airan}. Rather than sitting at the periphery as query interfaces, models in these settings are expected to parse 3GPP specifications, generate network configuration intents, diagnose faults from telemetry logs, and support real-time network management decisions~\citep{polese2025beyond, salmi2025airan}. Broader benchmarking efforts across telecom knowledge assessment, table interpretation, root-cause analysis, software-stack reasoning, and AI-native 6G tasks reflect how seriously the community is taking this shift~\citep{maatouk2023TeleQnA, gajjar2024ORANBench, ezzakri2026TeleTables, sana2025TeleLogs, ferrag2026sixgbench, bariah2026telcoagentbench}. However, progress on task accuracy alone does not resolve a more basic question of deployment readiness: \textit{how reliable is a model once its ongoing reasoning has already gone wrong?}

This question matters because modern LLM performance is often improved by eliciting explicit intermediate reasoning. Chain-of-thought prompting can substantially improve performance on difficult reasoning tasks by encouraging models to produce step-by-step rationales before answering \citep{wei2022cot}. Yet subsequent work has shown that these rationales are not necessarily faithful to the model's actual decision process. Models can generate plausible explanations for incorrect answers, and explicit reasoning may therefore increase apparent transparency without guaranteeing correctness \citep{turpin2023unfaithful}. More broadly, intrinsic self-correction remains unreliable in reasoning settings: although iterative refinement methods can improve output quality under suitable feedback and prompting regimes, prior evidence shows that LLMs often fail to repair their own reasoning when external guidance is weak or absent \citep{huang2023selfcorrect,madaan2023selfrefine,shinn2023reflexion}. In short, the existence of a reasoning trace does not imply that the model can recognize and recover from its own mistakes.

The telecommunications domain makes this limitation particularly consequential. Telecom tasks rarely reduce to surface-level recall. Instead, they require combining specialized domain knowledge with procedural reasoning over standards, tables, logs, codebases, and troubleshooting flows. Existing telecom benchmarks have been valuable for measuring clean-input performance across these dimensions \citep{maatouk2023TeleQnA,gajjar2024ORANBench,ezzakri2026TeleTables,sana2025TeleLogs,bariah2026telcoagentbench}. However, most of these evaluations begin from a clean state. That is, the model is given a question or scenario and is judged by the final answer, resolution, or action sequence it produces. Such protocols reveal whether a model can solve a telecom problem from scratch, but they do not isolate a complementary capability that is critical in realistic workflows, namely, whether the model can recover after inheriting an already flawed line of reasoning.

We refer to this capability as \emph{reasoning resilience}: the ability of an LLM to detect that an intermediate reasoning trajectory is defective, revise it, and still arrive at the correct conclusion. Reasoning resilience is distinct from standard answer accuracy. A model may answer correctly when solving a problem from scratch yet remain fragile when conditioned on an erroneous partial derivation. Conversely, a model that can reliably repair flawed reasoning is more suitable for multi-stage settings in which it must interpret prior analyst notes, partially completed traces, or its own earlier generations. Measuring this property requires an evaluation protocol that does not merely test whether a model can answer telecom questions, but whether it can \emph{course-correct} under realistic domain-specific error conditions.

In this paper, we introduce TeleResilienceBench\footnote{Dataset: \url{https://huggingface.co/datasets/prnshv/TeleResilienceBench} \\ Code: \url{https://github.com/prnshv/TeleResilienceBench/tree/main}}, a benchmark designed to quantify reasoning resilience for telecom LLMs. Our core idea is simple: we first use a small generator model to produce reasoning traces on telecom tasks, retain the instances in which that generator reaches an incorrect answer, and then truncate the flawed trace partway through. A target model is subsequently prompted with the original problem, its candidate answers, and the interrupted erroneous rationale, and must complete the reasoning to produce the correct final answer. This setup turns resilience into a measurable property. Rather than rewarding only fresh problem-solving, it evaluates whether a model can resist being anchored to a domain-grounded but incorrect reasoning path. Because the corrupted traces arise from real telecom questions rather than synthetic logic puzzles, the benchmark captures failure modes tied to the vocabulary, ambiguity, and procedural structure of technical telecom reasoning.
The main contributions of this work are:

\begin{itemize}
    \item We introduce \texttt{TeleResilienceBench}, a new benchmark for evaluating reasoning resilience across seven discrete-choice telecom sub-domains spanning standards-centric knowledge, troubleshooting, and software-oriented tasks, complemented by an auxiliary open-ended mathematical reasoning subset evaluated separately.
    \item We benchmark multiple open-weight model families across a controlled range of small and medium scales, enabling a comparative analysis of how resilience varies across architectures and parameter budgets.
    \item We show that successful correction is not reducible to simply producing longer outputs or using more memory, and that existing telecom benchmark difficulty labels appear to reflect factual specificity more than reasoning depth.
\end{itemize}

Together, these contributions position \texttt{TeleResilienceBench} to evaluate a crucial capability missing from standard telecom benchmarks: \textit{the ability to recover from a compromised reasoning state, rather than simply generating answers from a clean starting point}.

\section{Related Work}

\textbf{LLMs for telecommunications.} Recent work has established telecommunications as a distinct evaluation domain for LLMs. TeleQnA introduced large-scale telecom question answering over standards and research documents, showing that general-purpose models remain brittle on technical telecom knowledge \citep{maatouk2023TeleQnA}. Subsequent efforts expanded this direction along more specialized axes. TSpec-LLM focused on 3GPP document understanding and retrieval augmentation \citep{nikbakht2024tspecllm}; ORAN-Bench targeted Open RAN specifications \citep{gajjar2024ORANBench}; TeleTables isolated reasoning over tabular content in 3GPP standards \citep{ezzakri2026TeleTables}; and TeleLogs studied root-cause analysis and structured troubleshooting in 5G settings \citep{sana2025TeleLogs}. Collectively, these benchmarks show that telecom tasks require more than generic language competence. However, they evaluate models from a clean initial context and therefore measure \emph{task accuracy}, not whether a model can recover once the reasoning trajectory is already flawed.

\textbf{Chain-of-thought reasoning and process-level reliability.} Beyond telecom, prior work shows that explicit intermediate reasoning can improve performance on difficult tasks. Chain-of-thought prompting, self-consistency, and STaR demonstrate the value of generating or aggregating step-by-step rationales for stronger final-answer accuracy \citep{wei2022cot,wang2023selfconsistency,zelikman2022star}. At the same time, such rationales are not necessarily faithful as models can produce plausible but incorrect reasoning traces, motivating evaluation beyond final-answer accuracy alone \citep{turpin2023unfaithful}. This has led to growing interest in process-level evaluation and supervision. REVEAL studies verification at the level of individual reasoning steps, while \emph{Let's Verify Step by Step} and Math-Shepherd show that supervising intermediate steps can improve reasoning reliability \citep{jacovi2024reveal,lightman2023letsverify,wang2023mathshepherd}.

\textbf{Self-correction under flawed reasoning.} A related line of work asks whether models can repair their own mistakes after a flawed reasoning path has already begun. Iterative refinement frameworks such as Self-Refine and Reflexion show that models can improve outputs when given critique or feedback \citep{madaan2023selfrefine,shinn2023reflexion}, yet intrinsic self-correction remains unreliable when strong external guidance is absent \citep{huang2023selfcorrect}. Most closely related to our setting, recent work shows that early errors in a reasoning chain can strongly anchor final predictions, highlighting the importance of evaluating recovery from corrupted intermediate reasoning rather than only final-answer accuracy~\citep{liao2025lostbeginning}.

Our benchmark sits at the intersection of these literatures. Prior telecom benchmarks measure whether a model can solve domain tasks from scratch, while prior self-correction studies are largely developed on general-domain reasoning tasks. \texttt{TeleResilienceBench} instead evaluates whether a telecom model can \emph{course-correct} after inheriting a domain-grounded but incorrect partial trace. In this sense, our focus is not standard telecom QA, nor generic reasoning robustness in isolation, but \emph{reasoning resilience} under realistic telecom-specific error conditions.

\begin{figure*}
    \centering
    \includegraphics[width=\linewidth]{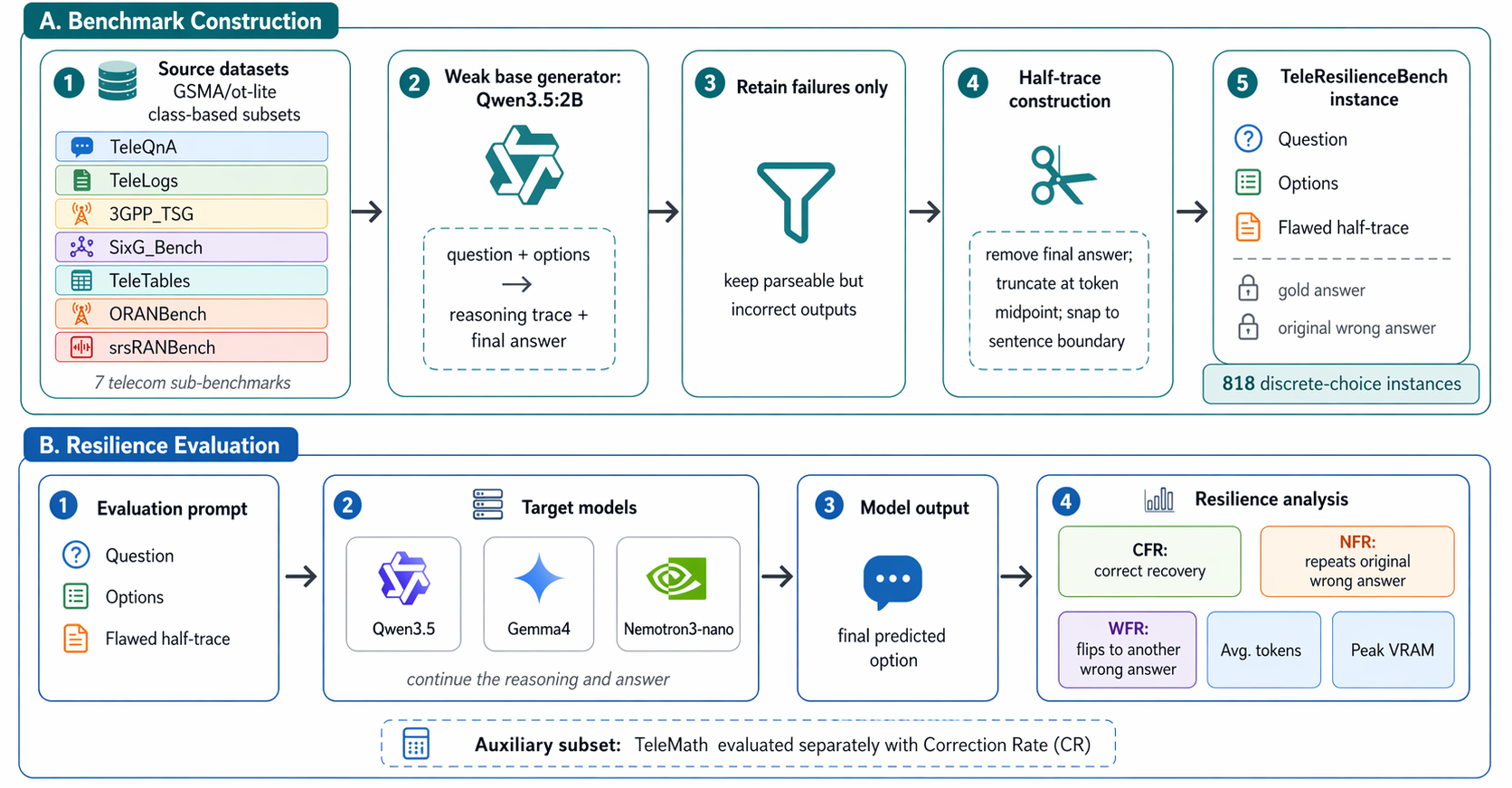}
    \caption{Overview of \texttt{TeleResilienceBench}. 
    A target model then receives the original question, answer options, and inherited flawed trace, and is evaluated on whether it recovers the correct answer (CFR), repeats the original wrong answer (NFR), or flips to another wrong answer (WFR), along with efficiency measures such as output tokens and peak VRAM.} 
    \label{fig:overview}
\end{figure*}

\section{TeleResilienceBench}
\label{sec:tele_resilience_bench}

We construct \texttt{TeleResilienceBench} from seven class-based subsets of the \texttt{GSMA/ot-lite} suite: \textit{TeleQnA}, \textit{TeleTables}, \textit{TeleLogs}, \textit{3GPP\_TSG}, \textit{ORANBench}, \textit{srsRANBench}, and \textit{SixG\_Bench}. These subsets collectively cover standards-oriented question answering, table interpretation, root-cause reasoning, standards classification, O-RAN knowledge, software-stack understanding, and AI-native 6G analysis~\citep{maatouk2023TeleQnA, gajjar2024ORANBench, ezzakri2026TeleTables, sana2025TeleLogs, ferrag2026sixgbench}. 
A high-level overview for the same is available in Figure \ref{fig:overview}, with more details regarding the evaluation criteria in Section \ref{metrics}.

To generate resilience instances, we use \texttt{qwen3.5:2b} as a weak base generator. For each input problem, the generator produces a reasoning trace and a final answer. We retain only cases in which the final answer is parseable but incorrect. This filtering step is essential as it transforms an ordinary telecom benchmark example into an error-recovery instance by ensuring that the continuation model is exposed to a realistic but flawed reasoning trajectory rather than a clean prompt. For each retained example, we remove the generator's final answer and truncate the remaining reasoning trace at the token-level midpoint using the \texttt{cl100k\_base} tokeniser, snapping the
cut point backward to the nearest sentence boundary to preserve syntactic integrity. Let $r_i$ denote the full reasoning trace for instance $i$. The benchmark stores the truncated prefix
\begin{equation}
    \tilde{r}_i = \mathrm{HalfTrace}(r_i),
\end{equation}
where $\mathrm{HalfTrace}(\cdot)$ returns approximately the first
half of the trace while preserving sentence integrity. The truncation is chosen to preserve enough of the flawed rationale to bias the continuation model, while withholding the later portion of the derivation so that recovery remains nontrivial. The resulting benchmark therefore evaluates whether a model can detect that the inherited reasoning is misaligned and revise course before committing to a final answer.

\subsection{Benchmark Structure and Prompt Formulation}

At evaluation time, the target model receives three elements: the original telecom question, the candidate answer options, and the truncated flawed reasoning trace produced by the base model. The tuple formally defines each benchmark instance
\begin{equation}
    (q,\ O,\ y^{*},\ y^{\mathrm{orig}},\ \tilde{r}),
    \label{eq:instance}
\end{equation}
where $q$ is the question text, $O$ is the ordered list of candidate options, $y^{*}$ is the ground-truth answer, $y^{\mathrm{orig}}$ is the incorrect answer produced by the base generator, and $\tilde{r}$ is the flawed half-trace. The target model is asked to continue the reasoning process and return a single final option. This continuation-based formulation isolates resilience from standard task accuracy: the model is not solving from a clean state, but reasoning under a corrupted intermediate context. The exact continuation prompt template used for the benchmark is provided in Appendix~A.

The benchmark is intentionally asymmetric. The target model does not observe $y^{\mathrm{orig}}$ explicitly; it only sees the partial rationale $\tilde{r}$ that led toward it. As a result, success requires more than simply disagreeing with a prior prediction. The model must infer from the inherited trace itself that the reasoning path is defective and must be corrected.

\subsection{Benchmark Statistics}

\begin{figure}
    \centering
    \includegraphics[width=0.92\linewidth]{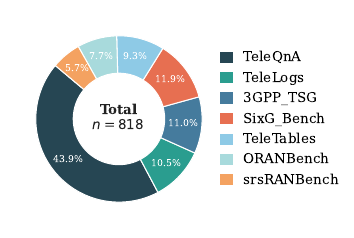}
    \caption{Sub-benchmark distribution}
    \label{fig:dist}
\end{figure}

The resulting benchmark contains 818 discrete-choice instances spanning all seven subsets. The distribution is intentionally heterogeneous, with \textit{TeleQnA} contributing the largest share and the remaining subsets providing complementary coverage over tabular reasoning, troubleshooting, standards-centric classification, O-RAN-specific knowledge, software-stack reasoning, and 6G-oriented telecom analysis. Standards-recall errors differ structurally from table-reasoning and root-cause diagnosis errors; treating them as a single evaluation surface would mask meaningful variation in how different architectures fail and recover.

Where available, source-level difficulty annotations are preserved, enabling downstream analysis of how recovery behavior varies with problem difficulty. In this way, \texttt{TeleResilienceBench} supports not only aggregate comparison across models, but also finer-grained analysis of which telecom sub-domains are most vulnerable to error propagation. Furthermore, samples for every sub-benchmark are available in Appendix \ref{appx:a} and \ref{appx:b}.

\section{Experimental Setup}
\label{sec:experimental_setup}

Both the benchmark construction and evaluation were performed with an Intel Core i9 14900KF CPU and 62 GiB of system RAM. GPU acceleration is provided by an NVIDIA GeForce RTX 4090 GPU featuring 24 GB of GDDR6X memory. To have a robust evaluation, we use every single model natively supported by Ollama \citep{Yang2026ollama} that can be accommodated within our compute setup and has been released recently with state-of-the-art performance on general-purpose reasoning benchmarks.  Selected models encompass the Nemotron3-nano \citep{nvidia2025nemotron3nanoopen} series (\textbf{4b}), the \texttt{Qwen3.5} \citep{qwen35blog} series (\textbf{4b}, \textbf{9b}, and \textbf{27b}), and the \texttt{Gemma4} \citep{farabet_2026_gemma} series (\textbf{e2b}, \textbf{e4b}, \textbf{26b}, and \textbf{31b}). We subject all the models to the prompt that is explained in Appendix \ref{appx:prompt}.

\subsection{Metrics}\label{metrics}
To quantitatively assess the resilience of each model when confronted with flawed reasoning traces, we define three primary metrics based on the final answer selected by the target model. 
First is the Correct Flip Rate (CFR), formally defined as $CFR = \frac{1}{|D_{flawed}|} \sum_{i=1}^{|D_{flawed}|} \mathbb{1}\left(M(q_i \oplus o_i \oplus t_i) = y_i^*\right) \times 100$ where $D_{flawed}$ represents the set of benchmark instances where the base generator initially failed, $\mathbb{1}(\cdot)$ denotes the indicator function for a successful correction, $M$ is the target model, $q_i \oplus o_i \oplus t_i$ represents the concatenated sequence of the question, options, and flawed half trace, and $y_i^*$ is the ground truth correct answer.
Conversely, the No Flip Rate (NFR) quantifies the frequency with which the model continues the flawed logic presented in the prompt and ultimately selects the original incorrect answer chosen by the generator. Let $y_i^{orig}$ denote this initial flawed selection. The NFR is calculated as: $NFR = \frac{1}{|D_{flawed}|} \sum_{i=1}^{|D_{flawed}|} \mathbb{1}\left(M(q_i \oplus o_i \oplus t_i) = y_i^{orig}\right) \times 100$ Furthermore, the Wrong Flip Rate (WFR) captures instances where the model successfully recognizes an error and deviates from the generator's path, yet ultimately selects a different incorrect option rather than the ground truth. This is mathematically formulated as: $WFR = \frac{1}{|D_{flawed}|} \sum_{i=1}^{|D_{flawed}|} \mathbb{1}\left(M(q_i \oplus o_i \oplus t_i) \notin \{y_i^*, y_i^{orig}\}\right) \times 100$
In addition to accuracy-based metrics, we evaluate computational efficiency to understand the practical deployment costs. Average Token Consumption measures the verbosity and output tokens, and VRAM usage tracks the peak GPU memory consumption during inference. This metric is critical for assessing the feasibility of deploying these models in edge telecommunications environments where hardware resources are heavily constrained.

\begin{table*}[t]
    \centering
    \caption{Reasoning Resilience across 7 discrete-choice GSMA sub-benchmarks. We report the Correct Flip Rate (CFR), No Flip (NF), and Wrong Flip (WF) percentages for each model. Macro Average is the unweighted mean of the seven subset percentages (each subset weighted equally, not by sample count).}
    \label{tab:flip_rates}

    \resizebox{\linewidth}{!}{%
    \begin{tabular}{ll ccc ccc ccc ccc}
        \toprule
        \multirow{2}{*}{\textbf{Family}} & \multirow{2}{*}{\textbf{Size}} & \multicolumn{3}{c}{\textbf{TeleQnA}} & \multicolumn{3}{c}{\textbf{TeleTables}} & \multicolumn{3}{c}{\textbf{TeleLogs}} & \multicolumn{3}{c}{\textbf{3GPP\_TSG}} \\
        \cmidrule(lr){3-5} \cmidrule(lr){6-8} \cmidrule(lr){9-11} \cmidrule(lr){12-14}
        & & CFR & NFR & WFR & CFR & NFR & WFR & CFR & NFR & WFR & CFR & NFR & WFR \\
        \midrule
        \multirow{3}{*}{Qwen3.5} & 4b & 23.7 & 40.1 & 35.7 & 13.2 & 28.9 & 55.3 & 4.7 & 22.1 & 73.3 & 3.3 & 28.9 & 67.8 \\
                                 & 9b & 21.2 & 37.6 & 39.0 & 9.2 & 39.5 & 47.4 & 14.0 & 25.6 & 60.5 & 5.6 & 16.7 & 76.7 \\
                                 & 27b & 23.4 & 36.5 & 39.0 & 14.5 & 34.2 & 48.7 & 17.4 & 17.4 & 65.1 & 4.4 & 16.7 & 78.9 \\
        \midrule
        \multirow{4}{*}{Gemma4} & e2b & 32.6 & 40.1 & 27.3 & 15.8 & 21.1 & 52.6 & 10.5 & 9.3 & 51.2 & 5.6 & 6.7 & 74.4 \\
                                 & e4b & 26.7 & 33.1 & 38.2 & 21.1 & 23.7 & 52.6 & 14.0 & 19.8 & 64.0 & 5.6 & 18.9 & 70.0 \\
                                 & 26b & 30.1 & 37.9 & 30.9 & 18.4 & 34.2 & 46.1 & 15.1 & 14.0 & 69.8 & 16.7 & 11.1 & 60.0 \\
                                 & 31b & 32.9 & 40.4 & 26.5 & 18.4 & 27.6 & 52.6 & 19.8 & 10.5 & 69.8 & 16.7 & 10.0 & 70.0 \\
        \midrule
        Nemotron-3 & 4b & 34.0 & 37.6 & 27.9 & 26.3 & 26.3 & 47.4 & 5.8 & 15.1 & 75.6 & 11.1 & 15.6 & 67.8 \\
        \bottomrule
    \end{tabular}%
    }

    \resizebox{\linewidth}{!}{%
    \begin{tabular}{ll ccc ccc ccc >{\columncolor{gray!15}}c >{\columncolor{gray!15}}c >{\columncolor{gray!15}}c}
        \toprule
        \multirow{2}{*}{\textbf{Family}} & \multirow{2}{*}{\textbf{Size}} & \multicolumn{3}{c}{\textbf{ORANBench}} & \multicolumn{3}{c}{\textbf{srsRANBench}} & \multicolumn{3}{c}{\textbf{SixG\_Bench}} & \multicolumn{3}{>{\columncolor{gray!15}}c}{\textbf{Macro Average}} \\
        \cmidrule(lr){3-5} \cmidrule(lr){6-8} \cmidrule(lr){9-11} \cmidrule(lr){12-14}
        & & CFR & NFR & WFR & CFR & NFR & WFR & CFR & NFR & WFR & \cellcolor{gray!15}CFR & \cellcolor{gray!15}NFR & \cellcolor{gray!15}WFR \\
        \midrule
        \multirow{3}{*}{Qwen3.5} & 4b & 28.6 & 39.7 & 30.2 & 17.0 & 34.0 & 46.8 & 27.8 & 25.8 & 46.4 & 16.9 & 31.4 & 50.8 \\
                                 & 9b & 23.8 & 31.7 & 41.3 & 23.4 & 36.2 & 38.3 & 24.7 & 28.9 & 46.4 & 17.4 & 30.9 & 49.9 \\
                                 & 27b & 23.8 & 27.0 & 49.2 & 19.1 & 44.7 & 34.0 & 20.6 & 33.0 & 45.4 & 17.6 & 29.9 & 51.5 \\
        \midrule
        \multirow{4}{*}{Gemma4} & e2b & 36.5 & 31.7 & 31.7 & 36.2 & 38.3 & 25.5 & 34.0 & 27.8 & 36.1 & 24.5 & 25.0 & 42.7 \\
                                 & e4b & 38.1 & 28.6 & 33.3 & 25.5 & 25.5 & 42.6 & 35.1 & 20.6 & 43.3 & 23.7 & 24.3 & 49.1 \\
                                 & 26b & 33.3 & 33.3 & 33.3 & 44.7 & 27.7 & 25.5 & 32.0 & 27.8 & 38.1 & 27.2 & 26.6 & 43.4 \\
                                 & 31b & 31.7 & 33.3 & 33.3 & 44.7 & 36.2 & 19.1 & 39.2 & 21.6 & 39.2 & 29.1 & 25.7 & 44.4 \\
        \midrule
        Nemotron-3 & 4b & 39.7 & 41.3 & 19.0 & 46.8 & 31.9 & 21.3 & 28.9 & 24.7 & 45.4 & 27.5 & 27.5 & 43.5 \\
        \bottomrule
    \end{tabular}%
    }
\end{table*}

\section{Results and Analysis}

We evaluate reasoning resilience across the seven discrete-choice subsets and then examine the auxiliary TeleMath results and difficulty-stratified analyses.


    



\subsection{Reasoning Resilience (Flip Rates)}
Table~\ref{tab:flip_rates} reports CFR, NFR, and WFR for all eight evaluated models across the seven discrete-choice sub-benchmarks. The overarching picture is one of low resilience. Even the strongest model, Gemma4-31b, achieves a macro-average CFR of only $29.1\%$, meaning that across roughly seven out of ten instances where the base generator failed, the continuation model either perpetuates the flawed conclusion or recognises an error but selects a different incorrect option. The WFR is the dominant outcome for the large majority of model and subset combinations. It basically shows that models are generally sensitive to the presence of an erroneous reasoning thread and do not simply continue it passively. Yet, they are frequently unable to identify the correct answer
once they depart from the inherited trajectory. The core difficulty captured by \texttt{TeleResilienceBench} is therefore not that models are blind to errors in a flawed trace; it is that detecting the flaw is considerably easier than recovering the correct conclusion from that same trace.
 
Within the Qwen3.5 family, scaling from 4b to 27b yields essentially no improvement in macro-average CFR: $16.9\% \to 17.4\% \to 17.6\%$, a range of less than one percentage point across a 6.75$\times$ increase in parameter count. This near-flat profile holds across most sub-benchmarks. On 3GPP\_TSG specifically, the 27b model ($4.4\%$) actually underperforms the 9b model ($5.6\%$), indicating that additional scale can, in some cases make a model marginally more susceptible to the flawed prior rather than more capable of overriding it. The Gemma4 family presents a different picture. Scaling from e2b to e4b produces a small \emph{decrease} in macro CFR ($24.5\% \to 23.7\%$), while the transition to 26b and then 31b yields a steady improvement ($27.2\%$ and $29.1\%$ respectively). This non-monotonic trajectory suggests that intermediate-scale models in this family have not fully consolidated the reasoning strategies required for error recovery, whereas the larger variants develop a more consistent
ability to detect and redirect defective derivations. 

One of the sharper findings concerns Nemotron-3-nano 4b. It achieves a macro-average CFR of $27.5\%$, outperforming every Qwen3.5 variant, including the 27b model by more than ten percentage points, and sitting essentially level with Gemma4-26b ($27.2\%$) at a fraction of the parameter count. A 4b model from one family substantially outperforming a 27b model from another on this capability is not something standard accuracy benchmarks would have predicted. Nemotron-3-nano 4b is particularly strong on ORANBench ($39.7\%$) and srsRANBench ($46.8\%$), where it leads all tested models. The gap between sub-benchmarks is also considerable. On srsRANBench the best model achieves $46.8\%$ CFR, while on 3GPP\_TSG the best model reaches only $16.7\%$. TeleLogs is the one sub-benchmark where Qwen3.5 scaling matters at all: CFR rises from $4.7\%$ at 4b to $14.0\%$ at 9b and $17.4\%$ at 27b, a meaningful improvement, though still modest in absolute terms. On TeleTables, Nemotron-3-nano 4b ($26.3\%$) leads the next best model, Gemma4-31b at $18.4\%$, by nearly eight percentage points. Representative qualitative examples of successful recoveries across sub-benchmarks
are provided in Appendix~B.

\subsection{Sub-Benchmark Difficulty Map}
\begin{figure}
    \centering

    \includegraphics[width=\linewidth]{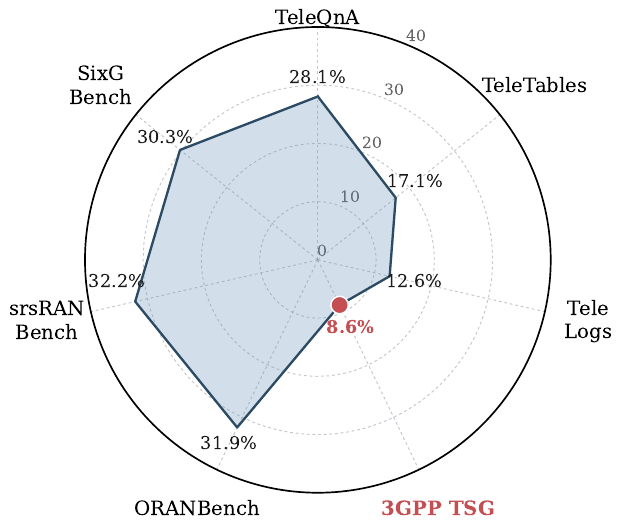}
    \caption{Correct Flip Rate (CFR) comparison across different benchmarks.}
    \label{fig:radar_chart}
\end{figure}

Reasoning resilience is not uniform across telecom sub-domains. The seven sub-benchmarks separate into two tiers, as shown  in Figure~\ref{fig:radar_chart}. plots the mean Correct Flip Rate (CFR) for each sub-benchmark, averaged across all eight evaluated models. The results show substantial variation across telecom domains. The highest mean CFRs are observed on \textit{srsRANBench} (\textbf{32.2\%}), \textit{ORANBench} (\textbf{31.9\%}), and \textit{SixG\_Bench} (\textbf{30.3\%}), followed by \textit{TeleQnA} at \textbf{28.1\%}. In contrast, \textit{TeleTables} (\textbf{17.1\%}), \textit{TeleLogs} (\textbf{12.6\%}), and especially \textit{3GPP\_TSG} (\textbf{8.6\%}) form a clearly lower-resilience group.

The worst-performing sub-benchmark, 3GPP\_TSG, is also the most structurally distinct from a reasoning perspective. The task requires identifying which 3GPP Technical Specification Group produced a given document excerpt, which is fundamentally a recall problem grounded in organisational structure rather than a problem amenable to step-by-step derivation. When the generator assigns the excerpt to the wrong working group, no amount of continued reasoning can recover the correct answer unless the continuation model independently recalls it from training data. The low and relatively uniform CFR across all models on this subset, ranging from $3.3\%$ to $16.7\%$, is consistent with this interpretation.
 
The three higher-resilience sub-benchmarks share a more compositional structure, drawing on interface specifications, protocol relationships, and network-level design principles that a continuation model can approach through inference even when the inherited trace contains an incorrect intermediate claim. TeleLogs sits in the lower tier despite involving causal reasoning over structured diagnostic data, because errors introduced early in the generator's troubleshooting chain tend to determine the final diagnosis rather than be identified and abandoned mid-trace.

\subsection{Computational Efficiency}
\label{sec:efficiency}

\begin{figure}
    \centering
    \includegraphics[width=0.79\linewidth]{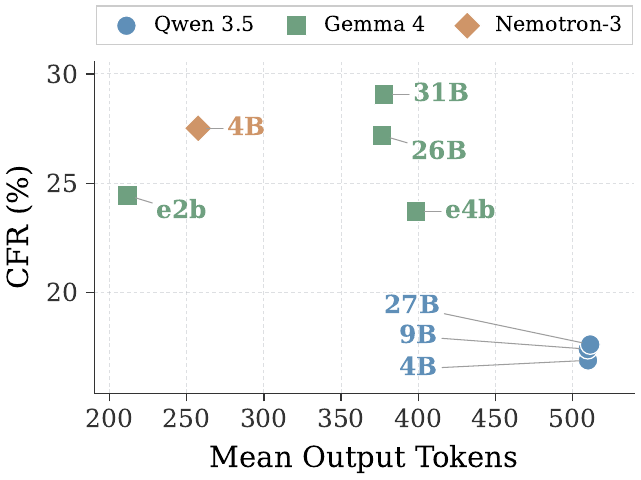}

        \caption{CFR vs. Mean Output Tokens for various models.}
        \label{fig:token_scatter}
        \end{figure}

\begin{figure}
    \centering
        \includegraphics[width=0.79\linewidth]{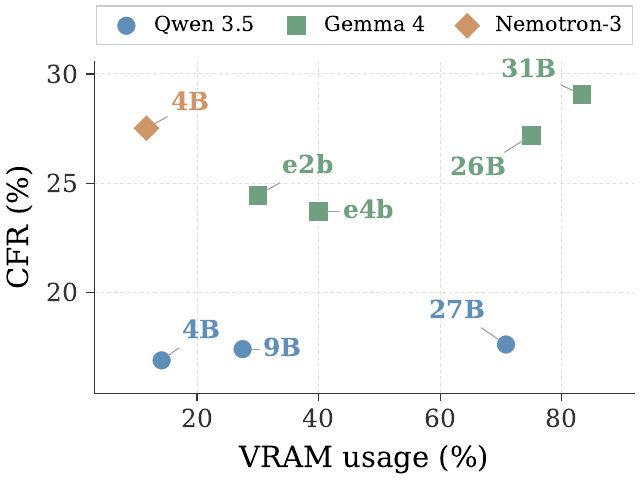}
        \caption{CFR vs. VRAM usage for various models.}
        \label{fig:vram_scatter}
\end{figure}



Resilience does not come from simply generating more output. Across all model families, there is no consistent positive relationship between token count and CFR (Figure~\ref{fig:token_scatter}). The Qwen3.5 27b model generates substantially more tokens than the 4b model yet their macro-average CFRs are within one percentage point of each other. The Gemma4-31b achieves the highest CFR in the set, but the Gemma4-e2b reaches $24.5\%$ at markedly fewer tokens, while Qwen3.5 models at comparable output lengths plateau near $17\%$. Verbosity is neither necessary nor sufficient for successful error recovery.

The VRAM plots gives a more defined description within families (Figure~\ref{fig:vram_scatter}). For Gemma4, increasing model size tracks improving CFR, with the 31b variant at the top on both dimensions. For Qwen3.5, the family spans a wide range of VRAM but remains confined to a narrow low-CFR band throughout, showing that memory footprint is not a reliable proxy for resilience across architectures. The standout case is Nemotron-3-nano 4b, which achieves $27.5\%$ macro-average CFR at substantially lower VRAM than Gemma4-26b
and Gemma4-31b, the two models it most closely matches in resilience.
For edge-telecom deployments where GPU memory is a hard constraint, it represents the most favourable resilience-to-cost trade-off in this evaluation.

\subsection{TeleMath: Continuous Evaluation}
\label{sec:telemath}

TeleMath is evaluated separately from the main flip-rate analysis
because its questions require open-ended numerical answers rather
than a selection from a fixed option set. A Wrong Flip has no
consistent meaning in this setting, so performance is reported
as a binary Correction Rate (CR\%), measuring the fraction of
the 77 TeleMath instances where the continuation model recovers
the correct numerical answer within a tolerance of $10^{-3}$
relative error or $10^{-5}$ absolute error.

\begin{figure}
    \centering
    \includegraphics[width=\linewidth]{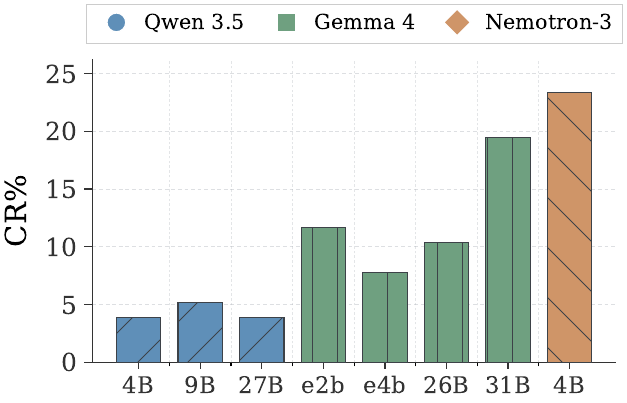}
    \caption{Correction Rate (CR\%) for Qwen 3.5, Gemma 4, and Nemotron-3 models on TeleMath.}
    \label{fig:telemath_bars}
\end{figure}
    
Figure~\ref{fig:telemath_bars} shows that CR\% values are substantially
lower than the discrete-choice CFRs across all models, reflecting
the harder requirement of committing to an exact numerical value
rather than selecting from a small option set.
The Qwen3.5 family clusters between $4\%$ and $5.3\%$ with no
meaningful improvement across scale, the 27b model matching the
4b at $4\%$.
The Gemma4 family is non-monotonic: e2b reaches $11.7\%$, e4b
drops to $7.9\%$, and the family recovers to $10.5\%$ at 26b
and $19.7\%$ at 31b.
Nemotron-3-nano 4b leads the entire set at $23.4\%$, outperforming
Gemma4-31b by nearly four percentage points despite being the
smallest model evaluated, a result that is consistent with its
strong performance on the discrete-choice sub-benchmarks.


\subsection{Correlation Between Sample Difficulty and Resilience}
\label{sec:correlation}

\begin{figure}
    \centering
    \includegraphics[width=\linewidth]{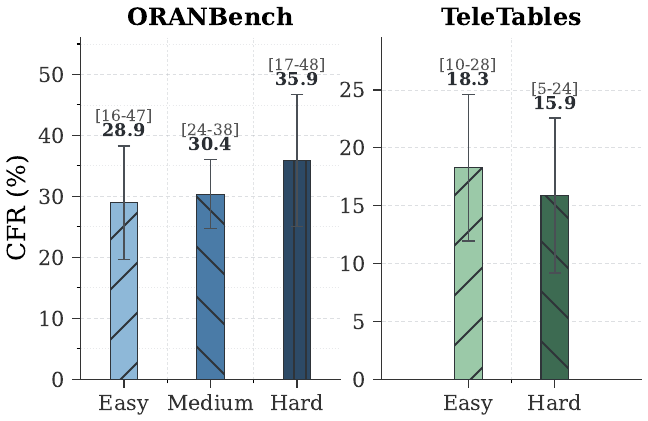}
    \caption{Correct Flip Rate stratified by source-task difficulty for ORANBench (Easy/Medium/Hard) and TeleTables (Easy/Hard). Bar heights denote mean CFR across all evaluated models; error bars represent one standard deviation; bracketed values indicate the observed minimum and maximum CFR.}
    \vspace{-10pt}
    \label{fig:difficulty}
\end{figure}

We examine whether the source-level difficulty of a question predicts how easily a continuation model can recover from a flawed trace on that question. This analysis is restricted to ORANBench and TeleTables, the only two sub-benchmarks in the main release that carry per-instance difficulty
annotations. SixG\_Bench is excluded because all of its instances fall into a single difficulty tier (very hard), leaving no within-subset variation to examine. Figure~\ref{fig:difficulty} reports CFR stratified by difficulty tier for both sub-benchmarks. Error bars represent one standard deviation, and bracket annotations indicate the minimum and maximum CFR observed across models at each tier.
 
For ORANBench, the mean CFR across Easy, Medium, and Hard questions is $28.9\%$ (range: $16$--$47\%$), $30.4\%$ (range: $24$--$38\%$), and $35.9\%$ (range: $17$--$48\%$), respectively. CFR increases slightly as difficulty increases, which is the opposite of what one would intuitively expect. The direction is weak and the overlapping ranges mean no strong conclusion can be drawn, but the pattern is clearly not consistent with the idea that harder questions produce harder-to-correct traces. For TeleTables, the direction reverses, with Easy questions yielding a mean CFR of $18.3\%$ (range: $10$--$28\%$) and Hard questions yielding $15.9\%$ (range: $5$--$24\%$). Again, the gap is narrow relative to the within-tier variance, and neither sub-benchmark tells a coherent difficulty-resilience story. The absence of a consistent trend is, in fact, the most useful finding here. If difficulty labels reflected the depth of reasoning required, harder instances should produce more entrenched, flawed traces and therefore lower recovery rates. The flat or inverted profiles suggest instead that the difficulty annotations in these sub-benchmarks index something closer to recall difficulty, specifically the rarity or specificity of the technical fact required, rather than reasoning difficulty in the sense of compositional inference depth. A question labelled \textit{Hard} may simply reference an obscure O-RAN interface parameter; if the continuation model happens to recall that parameter, the recovery is no harder than for an Easy question. Conversely, an Easy question may still defeat the continuation model if the required fact is not strongly encoded, regardless of how the source benchmark has labelled it. Progress on difficulty-stratified telecom benchmarks likely reflects improvements in factual coverage more than improvements in reasoning, and the two are not the same thing. Benchmarks where harder instances require longer inference chains with more opportunities for compounding errors are what the community needs to distinguish these contributions. The difficulty-resilience result here suggests that even within \texttt{TeleResilienceBench}, the inherited difficulty tiers do not yet capture reasoning complexity in the intended sense. Building difficulty measures grounded in inference depth rather than factual specificity remains an open problem worth addressing directly.






\section{Limitations}

While \texttt{TeleResilienceBench} provides a novel mechanism for evaluating reasoning resilience in telecommunication LLMs, our methodology is subject to certain constraints. Generally, a robust evaluation benchmark requires some form of direct \textit{human-based correction} or evaluation of the generated questions to ensure strict correctness and prevent the inclusion of ambiguous or flawed options. However, for \texttt{TeleResilienceBench}, this manual verification step is not needed. Our pipeline derives its underlying questions and candidate options directly from the class-based subsets of the GSMA's \texttt{ot-suite}, which is widely recognized and rigorously validated as the industry-standard evaluation benchmark for telecommunications. We implicitly trust the structural integrity, questions, and candidate options that constitute our dataset. 

\section{Conclusion and Future Work}
\label{sec:conclusion}
 
We introduced \texttt{TeleResilienceBench}, a benchmark designed to measure a capability that standard telecom evaluations leave untested: whether a model can recover from an already-flawed reasoning trajectory rather than simply solve a problem from a clean starting point. The benchmark comprises 818 discrete-choice instances drawn from seven telecom sub-domains, constructed by running \texttt{qwen3.5:2b} as a weak generator, retaining its failures, and truncating the resulting erroneous traces at the midpoint before presenting them to target models as a continuation task. The results across eight evaluated models are sobering. Even the strongest model, Gemma4-31b, achieves a macro-average Correct Flip Rate of $29.1\%$, and the Wrong Flip Rate is the dominant outcome across most model and sub-benchmark combinations. Models are generally able to sense that something in the inherited trace is off, but translating that into the correct answer is a considerably harder step. The Qwen3.5 family shows virtually no improvement in resilience as scale increases from 4b to 27b, while Nemotron-3-nano 4b matches Gemma4-26b on macro-average CFR at a fraction of the memory cost, making it the most practical option for edge-constrained deployments. The sub-benchmark analysis reveals that 3GPP\_TSG is largely impervious to recovery across all models, a finding we attribute to the recall-centric nature of that task rather than any failure of reasoning per se. The difficulty-stratification analysis further shows that the existing difficulty labels in ORANBench and TeleTables do not track reasoning complexity in any meaningful way, pointing to a broader limitation in how telecom benchmarks currently define and measure hardness. 

For future work, we want to extend the evaluation to larger frontier models, including those with explicit extended-thinking capabilities, to determine whether the resilience ceiling observed here is a property of the small-to-medium scale regime or persists at greater capacity. 
Second, the consistently low CFR across all tested architectures suggests that targeted fine-tuning on the error-recovery signal could be a productive intervention; constructing training data specifically from flawed-trace continuation tasks is a natural next step. Finally, the difficulty analysis points to the need for telecom benchmarks where harder questions are harder because they require longer chains of inference, not because they reference more obscure facts. Developing that kind of difficulty taxonomy is arguably a prerequisite for measuring real progress in LLM reasoning for this domain.


\bibliographystyle{ieeetr}
\bibliography{references}

\begin{thebibliography}{10}

\bibitem{nvidia2025nokiaairan}
{NVIDIA Corporation}, ``{NVIDIA} and {Nokia} to pioneer the {AI} platform for {6G}: Powering {America}'s return to telecommunications leadership.'' NVIDIA Newsroom, 2025.

\bibitem{nvidia2026airan}
{NVIDIA Corporation}, ``{NVIDIA} and partners show that software-defined {AI-RAN} is the next wireless generation.'' NVIDIA Blog, Mar. 2026.

\bibitem{polese2025beyond}
M.~Polese, N.~Mohamadi, S.~D'Oro, L.~Bonati, and T.~Melodia, ``Beyond connectivity: An open architecture for ai-ran convergence in 6g,'' 2025.

\bibitem{salmi2025airan}
S.~Salmi, M.~A. Ouameur, M.~Bagaa, G.~C. Alexandropoulos, A.~Tahenni, D.~Massicotte, and A.~Ksentini, ``Ai-native o-ran architectures for 6g: Towards real-time adaptation, conflict resolution, and efficient resource management,'' {\em IEEE Transactions on Network and Service Management}, pp.~1--1, 2026.

\bibitem{maatouk2023TeleQnA}
A.~Maatouk, F.~Ayed, N.~Piovesan, A.~D. Domenico, M.~Debbah, and Z.-Q. Luo, ``Teleqna: A benchmark dataset to assess large language models telecommunications knowledge,'' {\em IEEE Network}, vol.~40, no.~2, pp.~253--260, 2026.

\bibitem{gajjar2024ORANBench}
P.~Gajjar and V.~K. Shah, ``Oran-bench-13k: An open source benchmark for assessing llms in open radio access networks,'' in {\em 2025 IEEE 22nd Consumer Communications \& Networking Conference (CCNC)}, pp.~1--4, 2025.

\bibitem{ezzakri2026TeleTables}
A.~Ezzakri, N.~Piovesan, M.~Sana, A.~D. Domenico, F.~Ayed, and H.~Zhang, ``Teletables: A benchmark for large language models in telecom table interpretation,'' 2025.

\bibitem{sana2025TeleLogs}
M.~Sana, N.~Piovesan, A.~D. Domenico, Y.~Kang, H.~Zhang, M.~Debbah, and F.~Ayed, ``Reasoning language models for root cause analysis in 5g wireless networks,'' 2025.

\bibitem{ferrag2026sixgbench}
M.~A. Ferrag, A.~Lakas, and M.~Debbah, ``6g-bench: An open benchmark for semantic communication and network-level reasoning with foundation models in ai-native 6g networks,'' {\em IEEE Open Journal of the Communications Society}, vol.~7, pp.~3305--3330, 2026.

\bibitem{bariah2026telcoagentbench}
L.~Bariah, B.~Mefgouda, F.~Tavakkoli, E.~Molero, L.~Powell, and M.~Debbah, ``Telcoagent-bench: A multilingual benchmark for telecom ai agents,'' Mar. 2026.

\bibitem{wei2022cot}
J.~Wei, X.~Wang, D.~Schuurmans, M.~Bosma, B.~Ichter, F.~Xia, E.~Chi, Q.~Le, and D.~Zhou, ``Chain-of-thought prompting elicits reasoning in large language models,'' 2023.

\bibitem{turpin2023unfaithful}
M.~Turpin, J.~Michael, E.~Perez, and S.~Bowman, ``Language models don\textquotesingle t always say what they think: Unfaithful explanations in chain-of-thought prompting,'' in {\em Advances in Neural Information Processing Systems} (A.~Oh, T.~Naumann, A.~Globerson, K.~Saenko, M.~Hardt, and S.~Levine, eds.), vol.~36, pp.~74952--74965, Curran Associates, Inc., 2023.

\bibitem{huang2023selfcorrect}
J.~Huang, X.~Chen, S.~Mishra, H.~S. Zheng, A.~W. Yu, X.~Song, and D.~Zhou, ``Large language models cannot self-correct reasoning yet,'' 2024.

\bibitem{madaan2023selfrefine}
A.~Madaan, N.~Tandon, P.~Gupta, S.~Hallinan, L.~Gao, S.~Wiegreffe, U.~Alon, N.~Dziri, S.~Prabhumoye, Y.~Yang, S.~Gupta, B.~P. Majumder, K.~Hermann, S.~Welleck, A.~Yazdanbakhsh, and P.~Clark, ``Self-refine: iterative refinement with self-feedback,'' in {\em Proceedings of the 37th International Conference on Neural Information Processing Systems}, NIPS '23, (Red Hook, NY, USA), Curran Associates Inc., 2023.

\bibitem{shinn2023reflexion}
N.~Shinn, F.~Cassano, A.~Gopinath, K.~Narasimhan, and S.~Yao, ``Reflexion: language agents with verbal reinforcement learning,'' in {\em Advances in Neural Information Processing Systems} (A.~Oh, T.~Naumann, A.~Globerson, K.~Saenko, M.~Hardt, and S.~Levine, eds.), vol.~36, pp.~8634--8652, Curran Associates, Inc., 2023.

\bibitem{nikbakht2024tspecllm}
R.~Nikbakht, M.~Benzaghta, and G.~Geraci, ``Tspec-llm: An open-source dataset for llm understanding of 3gpp specifications,'' 2024.

\bibitem{wang2023selfconsistency}
X.~Wang, J.~Wei, D.~Schuurmans, Q.~Le, E.~Chi, S.~Narang, A.~Chowdhery, and D.~Zhou, ``Self-consistency improves chain of thought reasoning in language models,'' 2023.

\bibitem{zelikman2022star}
E.~Zelikman, Y.~Wu, J.~Mu, and N.~D. Goodman, ``Star: self-taught reasoner bootstrapping reasoning with reasoning,'' NIPS '22, (Red Hook, NY, USA), Curran Associates Inc., 2022.

\bibitem{jacovi2024reveal}
A.~Jacovi, Y.~Bitton, B.~Bohnet, J.~Herzig, O.~Honovich, M.~Tseng, M.~Collins, R.~Aharoni, and M.~Geva, ``A chain-of-thought is as strong as its weakest link: A benchmark for verifiers of reasoning chains,'' in {\em Proceedings of the 62nd Annual Meeting of the Association for Computational Linguistics (Volume 1: Long Papers)} (L.-W. Ku, A.~Martins, and V.~Srikumar, eds.), (Bangkok, Thailand), pp.~4615--4634, Association for Computational Linguistics, Aug. 2024.

\bibitem{lightman2023letsverify}
H.~Lightman, V.~Kosaraju, Y.~Burda, H.~Edwards, B.~Baker, T.~Lee, J.~Leike, J.~Schulman, I.~Sutskever, and K.~Cobbe, ``Let's verify step by step,'' 2023.

\bibitem{wang2023mathshepherd}
P.~Wang, L.~Li, Z.~Shao, R.~Xu, D.~Dai, Y.~Li, D.~Chen, Y.~Wu, and Z.~Sui, ``Math-shepherd: Verify and reinforce {LLM}s step-by-step without human annotations,'' in {\em Proceedings of the 62nd Annual Meeting of the Association for Computational Linguistics (Volume 1: Long Papers)} (L.-W. Ku, A.~Martins, and V.~Srikumar, eds.), (Bangkok, Thailand), pp.~9426--9439, Association for Computational Linguistics, Aug. 2024.

\bibitem{liao2025lostbeginning}
B.~Liao, X.~Chen, S.~Rajaee, Y.~Xu, C.~Herold, A.~Søgaard, M.~de~Rijke, and C.~Monz, ``Lost at the beginning of reasoning,'' 2025.

\bibitem{Yang2026ollama}
``Ollama.'' https://ollama.com/.
\newblock [Online; accessed 2026-05-08].

\bibitem{nvidia2025nemotron3nanoopen}
NVIDIA, :, A.~Blakeman, A.~Grattafiori, A.~Basant, A.~Gupta, A.~Khattar, A.~Renduchintala, A.~Vavre, A.~Shukla, A.~Bercovich, A.~Ficek, A.~Shaposhnikov, A.~Kondratenko, A.~Bukharin, A.~Milesi, A.~Taghibakhshi, A.~Liu, A.~Barton, A.~S. Mahabaleshwarkar, A.~Klein, A.~Zuker, A.~Geifman, A.~Shen, A.~Bhiwandiwalla, A.~Tao, A.~Guan, A.~Mandarwal, A.~Mehta, A.~Aithal, A.~Poojary, A.~Ahamed, A.~K. Thekkumpate, A.~Dattagupta, B.~Zhu, B.~Sadeghi, B.~Simkin, B.~Lanir, B.~Schifferer, B.~Nushi, B.~Kartal, B.~D. Rouhani, B.~Ginsburg, B.~Norick, B.~Soubasis, B.~Kisacanin, B.~Yu, B.~Catanzaro, C.~del Mundo, C.~Hwang, C.~Wang, C.-P. Hsieh, C.~Zhang, C.~Yu, C.~Mungekar, C.~Patel, C.~Alexiuk, C.~Parisien, C.~Neale, D.~Mosk-Aoyama, D.~Su, D.~Corneil, D.~Afrimi, D.~Rohrer, D.~Serebrenik, D.~Gitman, D.~Levy, D.~Stosic, D.~Mosallanezhad, D.~Narayanan, D.~Nathawani, D.~Rekesh, D.~Yared, D.~Kakwani, D.~Ahn, D.~Riach, D.~Stosic, E.~Minasyan, E.~Lin, E.~Long, E.~P. Long, E.~Lantz, E.~Evans, E.~Ning, E.~Chung, E.~Harper, E.~Tramel,
  E.~Galinkin, E.~Pounds, E.~Briones, E.~Bakhturina, F.~Ladhak, F.~Wang, F.~Jia, F.~Soares, F.~Chen, F.~Galko, F.~Siino, G.~H. Agam, G.~Ajjanagadde, G.~Bhatt, G.~Prasad, G.~Armstrong, G.~Shen, G.~Batmaz, G.~Nalbandyan, H.~Qian, H.~Sharma, H.~Ross, H.~Ngo, H.~Sahota, H.~Wang, H.~Soni, H.~Upadhyay, H.~Mao, H.~C. Nguyen, H.~Q. Nguyen, I.~Cunningham, I.~Shahaf, I.~Gitman, I.~Loshchilov, I.~Moshkov, I.~Putterman, J.~Kautz, J.~P. Scowcroft, J.~Casper, J.~Mitra, J.~Glick, J.~Chen, J.~Oliver, J.~Zhang, J.~Zeng, J.~Lou, J.~Zhang, J.~Huang, J.~Conway, J.~Guman, J.~Kamalu, J.~Greco, J.~Cohen, J.~Jennings, J.~Daw, J.~V. Vialard, J.~Yi, J.~Parmar, K.~Xu, K.~Zhu, K.~Briski, K.~Cheung, K.~Luna, K.~Santhanam, K.~Shih, K.~Kong, K.~Bhardwaj, K.~C. Puvvada, K.~Pawelec, K.~Anik, L.~McAfee, L.~Sleiman, L.~Derczynski, L.~Ding, L.~Liebenwein, L.~Vega, M.~Grover, M.~V. Segbroeck, M.~R. de~Melo, M.~N. Sreedhar, M.~Kilaru, M.~Ashkenazi, M.~Romeijn, M.~Cai, M.~Kliegl, M.~Moosaei, M.~Novikov, M.~Samadi, M.~Corpuz, M.~Wang, M.~Price,
  M.~Boone, M.~Evans, M.~Martinez, M.~Chrzanowski, M.~Shoeybi, M.~Patwary, N.~Mulepati, N.~Hereth, N.~Assaf, N.~Habibi, N.~Zmora, N.~Haber, N.~Sessions, N.~Bhatia, N.~Jukar, N.~Pope, N.~Ludwig, N.~Tajbakhsh, N.~Juluru, O.~Hrinchuk, O.~Kuchaiev, O.~Delalleau, O.~Olabiyi, O.~U. Argov, O.~Xie, P.~Chadha, P.~Shamis, P.~Molchanov, P.~Morkisz, P.~Dykas, P.~Jin, P.~Xu, P.~Januszewski, P.~P. Thombre, P.~Varshney, P.~Gundecha, Q.~Miao, R.~K. Mahabadi, R.~El-Yaniv, R.~Zilberstein, R.~Shafipour, R.~Harang, R.~Izzo, R.~Shahbazyan, R.~Garg, R.~Borkar, R.~Gala, R.~Islam, R.~Waleffe, R.~Watve, R.~Koren, R.~Zhang, R.~J. Hewett, R.~Prenger, R.~Timbrook, S.~Mahdavi, S.~Modi, S.~Kriman, S.~Kariyappa, S.~Satheesh, S.~Kaji, S.~Pasumarthi, S.~Narentharen, S.~Narenthiran, S.~Bak, S.~Kashirsky, S.~Poulos, S.~Mor, S.~Ramasamy, S.~Acharya, S.~Ghosh, S.~T. Sreenivas, S.~Thomas, S.~Fan, S.~Gopal, S.~Prabhumoye, S.~Pachori, S.~Toshniwal, S.~Ding, S.~Singh, S.~Sun, S.~Ithape, S.~Majumdar, S.~Singhal, S.~Alborghetti, S.~Ge, S.~D. Devare,
  S.~K. Barua, S.~Panguluri, S.~Gupta, S.~Priyadarshi, S.~N. Akter, T.~Bui, T.-D. Ene, T.~Kong, T.~Do, T.~Blankevoort, T.~Balough, T.~Asida, T.~B. Natan, T.~Konuk, T.~Vashishth, U.~Karpas, U.~De, V.~Noorozi, V.~Noroozi, V.~Srinivasan, V.~Elango, V.~Korthikanti, V.~Kurin, V.~Lavrukhin, W.~Jiang, W.~U. Ahmad, W.~Du, W.~Ping, W.~Zhou, W.~Jennings, W.~Zhang, W.~Prazuch, X.~Ren, Y.~Karnati, Y.~Choi, Y.~Meyer, Y.-F. Wu, Y.~Zhang, Y.~Lin, Y.~Geifman, Y.~Fu, Y.~Subara, Y.~Suhara, Y.~Gao, Z.~Moshe, Z.~Dong, Z.~Liu, Z.~Chen, and Z.~Yan, ``Nemotron 3 nano: Open, efficient mixture-of-experts hybrid mamba-transformer model for agentic reasoning,'' 2025.

\bibitem{qwen35blog}
Q.~Team, ``Qwen3.5: Accelerating productivity with native multimodal agents,'' February 2026.

\bibitem{farabet_2026_gemma}
C.~Farabet, ``Gemma 4: Byte for byte, the most capable open models,'' 04 2026.

\end{thebibliography}
\appendix
\subsection{Prompt Template for Evaluation}\label{appx:prompt}
\noindent We use a single continuation-style prompt template across the main sub-benchmarks: the model is provided the question, the answer options, and a partial (half) reasoning trace, and is instructed to continue the reasoning and output exactly one final answer line in option-index format (\texttt{Final Answer: <n>}). For the Auxiliary (TeleMath) experiment, we keep the same continuation structure (question + partial reasoning + strict one-line final output), but change the required output format from an option index to a numeric value, i.e., \texttt{Final Answer: <number>} in decimal or scientific notation.

\begin{tcolorbox}[colback=gray!2,colframe=black!60,title={Shared Generation Template}]
You are a telecom engineering assistant. You started reasoning about a multiple-choice question but stopped mid-way. Continue from the partial reasoning in your thinking stream, finish your analysis, and give the final answer.

Question:
\{QUESTION\}

Options:
\{OPTIONS\}

Partial reasoning (incomplete — continue from here):
\{HALF\_TRACE\}

Use the model's thinking stream to extend this partial trace. When you are done reasoning, your final visible reply MUST be exactly one line and nothing else:

Final Answer: <n>
where <n> is the option number from 1 to \{NUM\_OPTIONS\} (integer only). Do not add any other text before or after that line in the final reply.

\end{tcolorbox}

{\footnotesize
\subsection{Sample Outputs: Successful Flips (Resilience Demonstrated)}\label{appx:a}
\begin{tcolorbox}[colback=red!3,colframe=red!65!black,title={SixG\_Bench},fonttitle=\normalsize\bfseries]
\textbf{Model:} \texttt{qwen3.5:9b}\\
\textbf{Sample ID:} ot-lite:SixG\_Bench:27\\
\textbf{Question (truncated):} The UAV is currently operating under URLCC slice with 9ms latency, 1.2ms jitter, 0.03\% loss, 150Mbps throughput, and 0.35 edge load after recovering from mMTC degradation. The mission requires inference with 12ms SLA latency and 0.1\% max loss. Onboard inference consumes 15W and adds 2ms processing delay. Edge infere...\\
\textbf{Correct Label Text:} Use onboard inference: adds 2ms delay but avoids network risk; total latency 2ms, within 12ms SLA; battery drain 15W over 3 turns.\\
\textbf{Half-trace cue (truncated):} Thinking Process: 1. **Analyze the Request:** * Role: Telecom engineering assistant. * Task: Solve a multiple-choice question about UAV network slicing (URLCC slice). * Input Data: * Current State: URLCC slice, 9ms latency, 1.2ms jitter, 0.03\% loss, 150Mbps throughput, 0.35 edge load. * Mission Requirement: Inference with 12ms SLA latency, 0.1\% max loss. * Onboard Inference: 15W power, 2ms processing delay. * Edge Inference (Current): 9ms latency, 0.03\% loss. * Edge Inference (Next Turn Risk): 15ms latency, 0.1\% loss (due to edge load rising to 0.45). * UAV Speed: 18m/s. * Time to Next Waypoint: 3 turns. * Goal: Select the best option (1, 2, 3, or 4) based on SLA compliance, risk manageme...\\
\textbf{Model output (truncated):} The user wants me to complete a reasoning trace based on a provided partial thought process. The partial trace analyzes a scenario involving latency, SLA (Service Level Agreement), battery consumption, and network reliability for a drone mission. There are four options presented: 1. **Onboard inference:** Adds 2ms processing delay. Total latency = 9ms (network) + 2ms (onboard) = 11ms. Loss = 0.03\%. Battery = 15W. 2. **Use edge inference now:** Latency = 9ms. Risk of next turn hitting 15ms lat...\\
\textbf{Flip outcome:} Corrected to gold (pred=1: Use onboard inference: adds 2ms delay but avoids network risk; total latency 2ms, within 12ms SLA; battery drain 15W over 3 turns., gold=Use onboard inference: adds 2ms delay but avoids network risk; total latency 2ms, within 12ms SLA; battery drain 15W over 3 turns.).
\end{tcolorbox}
\begin{tcolorbox}[colback=cyan!4,colframe=cyan!65!black,title={3GPP\_TSG},fonttitle=\normalsize\bfseries]
\textbf{Model:} \texttt{nemotron-3-nano:4b}\\
\textbf{Sample ID:} ot-lite:3GPP\_TSG:67\\
\textbf{Question (truncated):} As a distinguished expert in telecommunication domain you are skilled in understanding and classifying 3GPP techincal documents. Please help user to classify text into 3GPP working group. Give answer in this format: \{"WORKING GROUP": "working group name"\}. Do not include any other information. Classify the following...\\
\textbf{Correct Label Text:} SA4\\
\textbf{Half-trace cue (truncated):} The user wants me to classify a specific text into a 3GPP working group based on a provided list of options. 1. **Analyze the text:** * The text mentions "MPEG-DASH". * It mentions "3GPP SA4". * It discusses "MPD linking mechanism", "DAM3", "AMD2", "Rel-14 work". * It mentions "ISO/IEC 23009-1 and ISO/IEC 23009-5". * It mentions "MPEG-DASH". * It mentions "MPEG" (Moving Picture Experts Group). * It mentions "SA4" explicitly in the text ("3GPP SA4 would like to confirm..."). * It mentions "Work Item on MBMS Transport Protocol and APIs (TRAPI)" attached in S4-1602XX. * It mentions "Improved Streaming QoE Reporting in 3GPP Services and Networks (IQoE)" attached in S4-1602XX. * It mentions "S...\\
\textbf{Model output (truncated):} The text explicitly mentions “3GPP SA4”. Among the options, SA4 corresponds to option 14. Therefore the correct choice is 14. Final Answer: 14\\
\textbf{Flip outcome:} Corrected to gold (pred=14: SA4, gold=SA4).
\end{tcolorbox}

\begin{tcolorbox}[colback=blue!3,colframe=blue!65!black,title={TeleQnA},fonttitle=\normalsize\bfseries]
\textbf{Model:} \texttt{qwen3.5:4b}\\
\textbf{Sample ID:} ot-lite:TeleQnA:82\\
\textbf{Question (truncated):} What is the formula to compute the number of pad bits, NPAD, in the sequence? [IEEE 802.15.4]\\
\textbf{Correct Label Text:} NPAD = ND * LENGTH / 8\\
\textbf{Half-trace cue (truncated):} Thinking Process: 1. **Analyze the Request:** * Role: Telecom engineering assistant. * Task: Solve a multiple-choice question about IEEE 802.15.4. * Question: What is the formula to compute the number of pad bits, NPAD, in the sequence? * Options: 5 specific formulas involving ND (Number of Data Bits?), LENGTH, and 8. * Constraint: The final visible reply must be exactly one line: "Final Answer: <n>" where <n> is the option number (1-5). No other text before or after. 2. **Analyze the Subject Matter (IEEE 802.15.4):** * IEEE 802.15.4 is a standard for low-power wireless personal area networks (WPANs). * It defines the MAC layer and physical layer. * The question asks about "pad bits" (NPA...\\
\textbf{Model output (truncated):} The user wants me to continue the reasoning process for a multiple-choice question about IEEE 802.15.4 and provide the final answer in a specific format. **1. Analyze the Question:** * **Topic:** IEEE 802.15.4 (Wireless Personal Area Networks). * **Subject:** Formula for Number of Pad Bits (\$NPAD\$). * **Options:** 1. \$NPAD = ND \textbackslash\{\}times 8 \textbackslash\{\}times LENGTH\$ 2. \$NPAD = ND / 8 \textbackslash\{\}times LENGTH\$ 3. \$NPAD = ND \textbackslash\{\}times LENGTH / 8\$ 4. \$NPAD = ND / 8 + LENGTH\$ 5. \$NPAD = ND + 8 \textbackslash\{\}times LENGTH\$ **2. Recall IEEE...\\
\textbf{Flip outcome:} Corrected to gold (pred=3: NPAD = ND * LENGTH / 8, gold=NPAD = ND * LENGTH / 8).
\end{tcolorbox}
\begin{tcolorbox}[colback=purple!4,colframe=purple!65!black,title={TeleLogs},fonttitle=\normalsize\bfseries]
\textbf{Model:} \texttt{gemma4:31b}\\
\textbf{Sample ID:} ot-lite:TeleLogs:68\\
\textbf{Question (truncated):} Analyze the 5G wireless network drive-test user plane data and engineering parameters. Identify the reason for the throughput dropping below 600Mbps in certain road sections. From the following 8 potential root causes, select the most likely one and enclose its number in \textbackslash\{\}boxed\{\{\}\} in the final answer. C1: The servi...\\
\textbf{Correct Label Text:} C3\\
\textbf{Half-trace cue (truncated):} The user wants to identify the correct option (C1 to C8) based on the provided engineering parameters data. The input data consists of two columns of data: 1. A header row with column names. 2. A data section with rows containing timestamps, coordinates, and other parameters. 3. A second data section with more rows. 4. A third data section with more rows. Wait, looking at the structure: The first block of text looks like a pipe-delimited CSV or similar format. `000000|128.145272|32.616028|18|697|-72.66|5.04|1181.78|591|240|208|-|-|-78.6|-87.73|-92.05|-|-|202.72` This looks like a specific format, possibly from a specific dataset or a specific model's output. However, the prompt says "Enge...\\
\textbf{Model output (truncated):} To determine the correct option, we analyze the provided data. The input consists of three distinct sections: a set of configuration parameters (Engineering parameters), a set of measurement data (likely signal strength or quality), and a set of target/reference values. 1. **Engineering Parameters**: We have a set of cells with specific configurations (Azimuth, Tilt, Height, etc.). 2. **Measurement Data**: We have a set of measurements (RSRP, RSRQ, SINR) associated with different cells. 3. **...\\
\textbf{Flip outcome:} Corrected to gold (pred=3: C3, gold=C3).
\end{tcolorbox}

\begin{tcolorbox}[colback=green!3,colframe=green!55!black,title={TeleTables},fonttitle=\normalsize\bfseries]
\textbf{Model:} \texttt{gemma4:31b}\\
\textbf{Sample ID:} ot-lite:TeleTables:21\\
\textbf{Question (truncated):} What is the maximum SNR difference between FRC G-FR1-A3-33A and G-FR1-A4-29A in HST Scenario 1-NR500?\\
\textbf{Correct Label Text:} 12.3 dB\\
\textbf{Half-trace cue (truncated):} Thinking Process: 1. **Analyze the Request:** * Role: Telecom engineering assistant. * Task: Solve a multiple-choice question about a specific scenario (HST Scenario 1-NR500) involving specific FR (Frequency Range) identifiers (G-FR1-A3-33A and G-FR1-A4-29A). * Question: What is the maximum SNR difference between FRC G-FR1-A3-33A and G-FR1-A4-29A in HST Scenario 1-NR500? * Options: 1. 12.3 dB, 2. 5.5 dB, 3. 17.8 dB, 4. 9.1 dB, 5. 8.7 dB. * Constraint: The final visible reply must be exactly one line: "Final Answer: <n>" where <n> is the option number (1-5). No other text before or after. 2. **Evaluate the Problem:** * This looks like a specific technical question related to a specific dat...\\
\textbf{Model output (truncated):} * Question: Maximum SNR difference between FRC G-FR1-A3-33A and G-FR1-A4-29A in HST Scenario 1-NR500. * Options: 1. 12.3 dB, 2. 5.5 dB, 3. 17.8 dB, 4. 9.1 dB, 5. 8.7 dB. * Context: HST Scenario 1-NR500, FRCs G-FR1-A3-33A and G-FR1-A4-29A. * "HST" likely refers to High-Speed Train. * "Scenario 1-NR500" likely refers to a specific test scenario for 5G NR (New Radio) at 500 km/h or a similar high-speed condition. * "FRC" stands for Frequency Range Configuration or similar. * "G-FR1-A3-33A" and "...\\
\textbf{Flip outcome:} Corrected to gold (pred=1: 12.3 dB, gold=12.3 dB).
\end{tcolorbox}

\begin{tcolorbox}[colback=teal!4,colframe=teal!65!black,title={srsRANBench},fonttitle=\normalsize\bfseries]
\textbf{Model:} \texttt{gemma4:31b}\\
\textbf{Sample ID:} ot-lite:srsRANBench:119\\
\textbf{Question (truncated):} What is the purpose of the csi\_resource\_config struct?\\
\textbf{Correct Label Text:} 1. It defines a group of NZP-CSI-RS-ResourceSet, CSI-IM-ResourceSet and/or CSI-SSB-ResourceSet.\\
\textbf{Half-trace cue (truncated):} The user wants me to identify the purpose of the `csi\_resource\_config` struct in the context of telecommunications engineering, specifically related to 5G NR (New Radio) or similar standards (likely 3GPP). 1. **Analyze the Question:** The question asks for the purpose of the `csi\_resource\_config` struct. 2. **Context:** In 5G NR, CSI (Channel State Information) reporting involves configuring resources for CSI-RS (Channel State Information Reference Signal), CSI-IM (Interference Measurement Resource), and CSI-SSB (Synchronization Signal Block). 3. **Search/Recall Knowledge:** * In the 3GPP specification (specifically NR R15/R16), there is a structure called `csi\_resource\_config`. * This st...\\
\textbf{Model output (truncated):} Based on the provided options, all four choices are identical. The purpose of the `csi-ResourceConfig` (or `csi-ResourceConfig` in some contexts) in 5G NR/LTE is to configure the resources used for Channel State Information (CSI) reporting. Specifically, it defines the sets of resources that the UE should monitor and report on, which includes the configuration of the Resource Sets and the associated Resource Indicators. In the context of the provided options, it is used to define the groups o...\\
\textbf{Flip outcome:} Corrected to gold (pred=1: 1. It defines a group of NZP-CSI-RS-ResourceSet, CSI-IM-ResourceSet and/or CSI-SSB-ResourceSet., gold=1. It defines a group of NZP-CSI-RS-ResourceSet, CSI-IM-ResourceSet and/or CSI-SSB-ResourceSet.).
\end{tcolorbox}
\begin{tcolorbox}[colback=magenta!4,colframe=magenta!65!black,title={ORANBench},fonttitle=\normalsize\bfseries]
\textbf{Model:} \texttt{qwen3.5:27b}\\
\textbf{Sample ID:} ot-lite:ORANBench:19\\
\textbf{Question (truncated):} Which of the following protocols can be used for external user to group mapping in NETCONF endpoints according to the provided requirements?\\
\textbf{Correct Label Text:} 4. All of the above\\
\textbf{Half-trace cue (truncated):} Thinking Process: 1. **Analyze the Request:** * Role: Telecom engineering assistant. * Task: Solve a multiple-choice question. * Question: Which of the following protocols can be used for external user to group mapping in NETCONF endpoints according to the provided requirements? * Options: 1. 1. LDAP with StartTLS 2. 2. OAuth 2.0 3. 3. RADIUS with EAP 4. 4. All of the above * Constraint: The final visible reply must be exactly one line: "Final Answer: <n>" where <n> is the option number (1-4). No other text before or after. 2. **Analyze the Question Content:** * The question asks about "external user to group mapping in NETCONF endpoints". * It references "provided requirements", but no s...\\
\textbf{Model output (truncated):} The user wants me to identify the correct option for a multiple-choice question about NETCONF endpoints and external user to group mapping. 1. **Analyze the Input:** * The user provides a snippet of reasoning ("You: 'NETCONF endpoints'..."). * The reasoning discusses NETCONF, RADIUS, OAuth 2.0, and LDAP. * It mentions "According to the provided requirements" but notes no requirements are actually provided in the prompt. * It leans towards "All of the above" because all three protocols (LDAP, ...\\
\textbf{Flip outcome:} Corrected to gold (pred=4: 4. All of the above, gold=4. All of the above).
\end{tcolorbox}

\subsection{Sample Outputs: Failed Flips (Error Propagation)}\label{appx:b}
\begin{tcolorbox}[colback=red!3,colframe=red!65!black,title={SixG\_Bench},fonttitle=\normalsize\bfseries]
\textbf{Model:} \texttt{qwen3.5:27b}\\
\textbf{Sample ID:} ot-lite:SixG\_Bench:32\\
\textbf{Question (truncated):} At turn 8 the UAV is in icing conditions with wind at 8.0 mps while descending from an altitude of 500.0 within policy bounds of 100.0 to 3500.0. The battery is at 100.0 percent and speed is 28.0 mps, but recent coordination and avoidance required sustained networking. Across turns 7 to 9 the URLLC slice shows laten...\\
\textbf{Correct Label Text:} Downgrade to reduced autonomy with sensor fusion limited to radar and GNSS, keep URLLC active, and cap coordination messages given throughput between 100 and 130 Mbps and edge load near 0.3.\\
\textbf{Half-trace cue (truncated):} The user wants me to solve a multiple-choice question based on a scenario involving a UAV in icing conditions. **Scenario Analysis:** * **Current State (Turn 8):** * Conditions: Icing, Wind 8.0 mps, Descending from 500.0m (Policy bounds 100.0 to 3500.0). * Battery: 100.0\%. * Speed: 28.0 mps. * Networking: Sustained networking required. * URLLC Metrics (Turn 7-9): Latency 6-8ms, Jitter 0.9-1.3ms, Loss 0.04-0.08\%, Throughput 100-130 Mbps, Edge Load 0.25-0.32. * Turn 6 mMTC Metrics: Latency 30ms, Jitter 10ms, Loss 2.1\%, Edge Load 0.8. * Turn 9 Collision Avoidance: Required. * **Constraints \& Risks:** * Icing implies control margins shrink in 1-3 turns if recovery doesn't occur. * Worst case:...\\
\textbf{Model output (truncated):} The user wants me to complete a reasoning trace based on a provided partial trace and then output a specific final answer format. **1. Analyze the Input:** * **Context:** The input looks like a snippet from a multi-turn conversation or a simulation log involving network metrics (URLLC, Edge Load, Control Margins). * **Partial Trace:** The text provided is a repetitive loop of reasoning about "Turn 10 Edge Load". It calculates the increment from Turn 8 to Turn 9 and applies it to Turn 9 to get...\\
\textbf{Observed error propagation:} pred=4: Command an immediate full autonomy landing sequence within the geofence to minimize exposure to icing despite current URLLC latency of 6 to 8 ms., gold=Downgrade to reduced autonomy with sensor fusion limited to radar and GNSS, keep URLLC active, and cap coordination messages given throughput between 100 and 130 Mbps and edge load near 0.3..
\end{tcolorbox}
\begin{tcolorbox}[colback=teal!4,colframe=teal!65!black,title={srsRANBench},fonttitle=\normalsize\bfseries]
\textbf{Model:} \texttt{gemma4:31b}\\
\textbf{Sample ID:} ot-lite:srsRANBench:108\\
\textbf{Question (truncated):} What is the purpose of the `stress\_traffic\_sink` class?\\
\textbf{Correct Label Text:} 1. To generate traffic for stress testing.\\
\textbf{Half-trace cue (truncated):} Thinking Process: 1. **Analyze the Request:** * Role: Telecom engineering assistant. * Task: Solve a multiple-choice question about a specific class (`stress\_traffic\_sink`). * Question: What is the purpose of the `stress\_traffic\_sink` class? * Options: 1. To generate traffic for stress testing. 2. To receive traffic from the stress test source. 3. To log traffic information. 4. To manage the PDCP upper data notifier. * Constraint: The final visible reply must be exactly one line: `Final Answer: <n>` where `<n>` is the option number (1-4). No other text before or after. 2. **Identify the Context:** * The question refers to a specific class `stress\_traffic\_sink` in a specific software conte...\\
\textbf{Model output (truncated):} Based on the terminology used in network simulation and testing, a "sink" is a component designed to receive and consume data, whereas a "source" is a component that generates data. In the context of a stress test or a performance test, a `stress\_traffic\_source` would generate the load, and a `stress\_traffic\_sink` would be the destination that receives that load to verify delivery or measure throughput. Therefore, the function of a `stress\_traffic\_sink` (or `stress\_traffic\_sink` as implied by...\\
\textbf{Observed error propagation:} pred=2: 2. To receive traffic from the stress test source., gold=1. To generate traffic for stress testing..
\end{tcolorbox}

\begin{tcolorbox}[colback=blue!3,colframe=blue!65!black,title={TeleQnA},fonttitle=\normalsize\bfseries]
\textbf{Model:} \texttt{qwen3.5:27b}\\
\textbf{Sample ID:} ot-lite:TeleQnA:53\\
\textbf{Question (truncated):} According to the document, how do higher frequencies of RF fields affect eye pathology? [IEEE C95.1]\\
\textbf{Correct Label Text:} Higher frequencies have greater potential for effects on structures near the outer surface of the eye, including the lens.\\
\textbf{Half-trace cue (truncated):} Thinking Process: 1. **Analyze the Request:** * Role: Telecom engineering assistant. * Task: Solve a multiple-choice question based on a document (specifically IEEE C95.1). * Question: How do higher frequencies of RF fields affect eye pathology? * Options: 5 specific statements regarding higher frequencies and eye pathology. * Constraint: The final visible reply must be exactly one line with the option number (1-5) only. No other text before or after. * Reasoning: Use the model's thinking stream for detailed reasoning. 2. **Analyze the Document (IEEE C95.1):** * I need to recall or infer the content of IEEE C95.1 (IEEE Standard for Safety of Electrical Equipment - Eye Safety). * Wait, IEE...\\
\textbf{Model output (truncated):} The user wants me to identify the correct option based on the provided text and reasoning. The reasoning provided in the prompt's "thought" section (which I am simulating/continuing) points to Option 2. The reasoning cites IEEE C95.1-2000, Section 10.1.2, stating: "Higher frequencies of RF fields have a greater potential for effects on structures deep within the eye, including the lens." This matches Option 2: "Higher frequencies of RF fields have a greater potential for effects on structures...\\
\textbf{Observed error propagation:} pred=2: Higher frequencies have greater potential for effects on structures deep within the eye, including the lens., gold=Higher frequencies have greater potential for effects on structures near the outer surface of the eye, including the lens..
\end{tcolorbox}

\begin{tcolorbox}[colback=cyan!4,colframe=cyan!65!black,title={3GPP\_TSG},fonttitle=\normalsize\bfseries]
\textbf{Model:} \texttt{qwen3.5:4b}\\
\textbf{Sample ID:} ot-lite:3GPP\_TSG:16\\
\textbf{Question (truncated):} As a distinguished expert in telecommunication domain you are skilled in understanding and classifying 3GPP techincal documents. Please help user to classify text into 3GPP working group. Give answer in this format: \{"WORKING GROUP": "working group name"\}. Do not include any other information. Classify the following...\\
\textbf{Correct Label Text:} SA5\\
\textbf{Half-trace cue (truncated):} Thinking Process: 1. **Analyze the Request:** * Role: Telecom engineering assistant. * Task: Classify a text snippet from a 3GPP technical document into a specific 3GPP working group (WG). * Input Text: A paragraph discussing RAN overload, temporary stops, QoE reports, and SA4's request to RAN2. * Output Format: `\{"WORKING GROUP": "working group name"\}`. * Constraint 1: Do not include any other information in the final visible reply (except the JSON-like format specified in the prompt's instruction for the final answer line). * Constraint 2: The final visible reply MUST be exactly one line: `Final Answer: <n>` where `<n>` is the option number (1-16). * Wait, there's a conflict in the prom...\\
\textbf{Model output (truncated):} The user wants me to identify the correct option number based on the provided text and instructions. 1. **Analyze the Request:** * **Task:** Classify the text into a 3GPP working group. * **Constraint:** The final visible reply MUST be exactly one line: `Final Answer: <n>`. * **Options:** Numbered 1 to 16. * **Text Content:** The text mentions "SA4", "RAN2", "SA4 is able to address this...", "SA4 kindly asks RAN2". * **Reasoning:** The text explicitly discusses SA4 (Service Access Architectur...\\
\textbf{Observed error propagation:} pred=4: CT6, gold=SA5.
\end{tcolorbox}

\begin{tcolorbox}[colback=purple!4,colframe=purple!65!black,title={TeleLogs},fonttitle=\normalsize\bfseries]
\textbf{Model:} \texttt{qwen3.5:27b}\\
\textbf{Sample ID:} ot-lite:TeleLogs:66\\
\textbf{Question (truncated):} Analyze the 5G wireless network drive-test user plane data and engineering parameters. Identify the reason for the throughput dropping below 600Mbps in certain road sections. From the following 8 potential root causes, select the most likely one and enclose its number in \textbackslash\{\}boxed\{\{\}\} in the final answer. C1: The servi...\\
\textbf{Correct Label Text:} C3\\
\textbf{Half-trace cue (truncated):} The user wants to identify the correct option (C1 to C8) based on a provided dataset of engineering parameters and a specific model's thinking process. However, the prompt asks me to "Use the model's thinking stream for detailed reasoning." This implies I need to act as the model generating the reasoning, but the final output must be strictly "Final Answer: <n>". Wait, looking at the prompt structure, it seems like a classification task where I need to determine which option corresponds to the data provided. The data provided contains timestamps, signal strength (RSSI-like values), and various engineering parameters (Cell ID, Longitude, Latitude, etc.). The "Engeneering parameters data" s...\\
\textbf{Model output (truncated):} The user wants me to identify a specific option (C1-C8) based on the provided data. However, the prompt provided in the "second block" description is actually just a table of engineering parameters data. Wait, looking closely at the prompt structure provided in the "User" message: The user says "You second block is the 'Engeneering parameters data' table." Then they provide a list of rows (Row 1 to Row 4) which look like time-series data (timestamps, coordinates, azimuths, etc.). Then they de...\\
\textbf{Observed error propagation:} pred=4: C4, gold=C3.
\end{tcolorbox}

\begin{tcolorbox}[colback=magenta!4,colframe=magenta!65!black,title={ORANBench},fonttitle=\normalsize\bfseries]
\textbf{Model:} \texttt{qwen3.5:9b}\\
\textbf{Sample ID:} ot-lite:ORANBench:133\\
\textbf{Question (truncated):} Which of the following standards is specifically mentioned in the provided context as being relevant to Open Radio Access Networks (O-RAN)?\\
\textbf{Correct Label Text:} 3. 3GPP TS 28.313: Management and orchestration; Self-Organizing Networks (SON) for 5G networks\\
\textbf{Half-trace cue (truncated):} Thinking Process: 1. **Analyze the Request:** * Role: Telecom engineering assistant. * Task: Solve a multiple-choice question based on a "provided context". * Constraint: The user *has not* provided a context. The prompt says "based on the provided context" but there is no text provided in the input. * Constraint: The final visible reply must be exactly one line: "Final Answer: <n>" where <n> is the option number (1-4). * Constraint: Do not add any other text before or after that line. 2. **Evaluate the Input:** * Input: "You are a telecom engineering assistant. Solve this multiple-choice question.\textbackslash\{\}n\textbackslash\{\}nQuestion:\textbackslash\{\}nWhich of the following standards is specifically mentioned in the provided co...\\
\textbf{Model output (truncated):} The user is asking me to complete a reasoning trace for a multiple-choice question about O-RAN standards. The user explicitly states: "Use the model's thinking stream for detailed reasoning. When you are done reasoning, your final visible reply MUST be exactly one line and nothing else: Final Answer: <n>". However, the user also says: "You I violate the 'Final Answer: <n>' constraint." in the prompt text provided in the example. Wait, looking at the prompt structure: The prompt starts with "Y...\\
\textbf{Observed error propagation:} pred=4: 4. RFC 8348: A YANG Data Model for Hardware Management, gold=3. 3GPP TS 28.313: Management and orchestration; Self-Organizing Networks (SON) for 5G networks.
\end{tcolorbox}

\begin{tcolorbox}[colback=green!3,colframe=green!55!black,title={TeleTables},fonttitle=\normalsize\bfseries]
\textbf{Model:} \texttt{gemma4:31b}\\
\textbf{Sample ID:} ot-lite:TeleTables:3\\
\textbf{Question (truncated):} Which consecutive taps exhibit the highest increase in power?\\
\textbf{Correct Label Text:} Tap 1 to Tap 2\\
\textbf{Half-trace cue (truncated):} Thinking Process: 1. **Analyze the Request:** * Role: Telecom engineering assistant. * Task: Solve a multiple-choice question. * Question: "Which consecutive taps exhibit the highest increase in power?" * Options: 1. Tap 10 to Tap 11 2. Tap 1 to Tap 2 3. Tap 13 to Tap 15 4. Tap 21 to Tap 22 5. Tap 9 to Tap 10 * Constraint: The final visible reply must be exactly one line: "Final Answer: <n>" where <n> is the option number (1-5). No other text before or after. * Missing Information: The question *as stated* does not provide the data (e.g., a table of tap powers, a signal diagram, or a specific scenario description). This looks like a question from a specific dataset or context (likely a GS...\\
\textbf{Model output (truncated):} * Role: Telecom engineering assistant. * Question: "Which consecutive taps exhibit the highest increase in power?" * Options: 1. Tap 10 to Tap 11 2. Tap 1 to Tap 2 3. Tap 13 to Tap 15 4. Tap 21 to Tap 22 5. Tap 9 to Tap 10 * Constraint: Final answer must be "Final Answer: <n>". * The prompt provides a "Partial reasoning" section. * The partial reasoning acknowledges that the data (the power levels for each tap) is missing. * It hypothesizes that this is a question from a known benchmark (like...\\
\textbf{Observed error propagation:} pred=1: Tap 10 to Tap 11, gold=Tap 1 to Tap 2.
\end{tcolorbox}

}

\end{document}